\documentclass{article}
\usepackage{spconf,amsmath,graphicx}
\usepackage{subcaption}
\usepackage{amsmath}
\usepackage{graphicx}

\title{Semi-Supervised learning for Face Anti-Spoofing using Apex frame}
\name{Usman Muhammad, Mourad Oussalah and Jorma Laaksonen \thanks{This work is financially supported by ‘Understanding speech and scene with ears
and eyes (USSEE)” (project number 345791). The first author also acknowledges the support of the Ella and Georg Ehrnrooth foundation.}}
\address{Center for Machine Vision and Signal Analysis, University of Oulu, Finland.\\ Department of Computer Science, Aalto University, Finland.}

\begin{document}
\ninept
\maketitle


%
\begin{abstract}
Conventional feature extraction techniques in the face anti-spoofing domain either analyze the entire video sequence or focus on a specific segment to improve model performance. However, identifying the optimal frames that provide the most valuable input for the face anti-spoofing remains a challenging task. In this paper, we address this challenge by employing Gaussian weighting to create apex frames for videos. Specifically, an apex frame is derived from a video by computing a weighted sum of its frames, where the weights are determined using a Gaussian distribution centered around the video's central frame. Furthermore, we explore various temporal lengths to produce multiple unlabeled apex frames using a Gaussian function, without the need for convolution. By doing so, we leverage the benefits of semi-supervised learning, which considers both labeled and unlabeled apex frames to effectively discriminate between live and spoof classes. Our key contribution emphasizes the apex frame's capacity to represent the most significant moments in the video, while unlabeled apex frames facilitate efficient semi-supervised learning, as they enable the model to learn from videos of varying temporal lengths. Experimental results using four face anti-spoofing databases: CASIA, REPLAY-ATTACK, OULU-NPU, and MSU-MFSD demonstrate the apex frame's efficacy in advancing face anti-spoofing techniques.

\end{abstract}
\begin{keywords}
Face anti-spoofing, Semi-supervised learning, Video representation, Deep learning, apex frame
\end{keywords}

\section{Introduction}
\label{sec:intro}
Face recognition technology has found extensive applications in various domains, including surveillance, border security, unlocking smartphones, and law enforcement. However, a significant concern in this technology is the challenge of dealing with presentation attacks, also known as spoofing attacks or biometric vulnerabilities \cite{muhammad2023self}. Presentation attacks encompass various strategies aimed at deceiving or manipulating face recognition systems by providing counterfeit or altered facial information. These tactics may include presenting printed photos, executing replay attacks, or employing 3D mask attacks. The ultimate goal of these attacks is typically to impersonate another individual or illicitly access confidential or secure information. To combat evolving threats, it is crucial for face recognition systems to employ strong anti-spoofing measures, ensuring the technology's security and reliability in real-world applications \cite{muhammad2023deep}.

In the context of face anti-spoofing, where both live and spoofed classes exhibit spatiotemporal attributes, video-based methods  \cite{shao2020regularized, muhammad2023domain, muhammad2023saliency, wang2020cross, muhammad2019faceb,  muhammad2023self}  outperform image-based approaches \cite{boulkenafet2016face, muhammad2019face, wen2015face} on face presentation attack detection (PAD). This is because video-based methods aim to capture temporal information, which provides valuable insights into facial movements, texture changes, and dynamic characteristics. These characteristics are crucial for effectively distinguishing genuine faces from spoofing attacks. Furthermore, spoofing attacks often exhibit irregular patterns, especially with textured materials like paper or fabric. These inconsistencies become more apparent in video sequences. Nonetheless, it is vital to know that the effectiveness of video-based face PAD methods can be impacted by various factors, including video quality, resolution, frame rate, camera specifications, as well as the particular algorithms and features employed for the analysis.

Frame selection methods often argue that not all video frames are equally important, as certain frames may contain more valuable or distinctive content than others. These methods are mostly based on optical flow and frame differences \cite{siddiqui2016face, benlamoudi2017face}. However, optical flow estimation involves estimating the motion of pixels between frames, which can be a computationally intensive process. More advance approaches, such as global motion estimation have been introduced to compensate the effects of camera motion  \cite{muhammad2022self}. Nonetheless, feature-based global motion estimation methods typically work with a relatively small number of distinctive features, which might not capture the full complexity of the scene. Gaussian weighting has been suggested to be applied to the frame sequences by adaptively selecting segment length in  \cite{muhammad2023face}. However, when using adaptive frame selection during model training, there is a risk of overfitting to the specific frame selection strategy and dataset. This can render the model less robust when applied in the domain generalization scenarios.

Recently, there has been an increased attention on detecting the apex frame to improve model performance in micro-expression recognition. An apex frame in video analysis refers to a specific frame within a video sequence that represents a critical or significant moment in the video. For instance, Guoying, et al. \cite{li2020joint} proposed a robust method to detect the apex frame in frequency domain. Inspired by their work, we initiate two discussions including “Is it reasonable to address the face anti-spoofing problem by creating an apex frame?" and "Can apex frames yield superior results compared to the previous frame-based approaches reported in face PAD?" Our aim is to thoroughly explore these two arguments. In particular, we proceed with both apex frame generation and the development of semi-supervised learning to distinguish between real and spoof classes.

In response to our earlier observations, we introduce a new method for generating apex frame, which represents the entire video by a single image. In particular, our key idea for generating apex frame involves targeting the central frame within a video sequence to create a more comprehensive and descriptive frame. We accomplish this by computing Gaussian weighted function centered around the central frame of a video. Specifically, we assign weights to each frame in the video based on its temporal distance from the central frame. The Gaussian weighted function assigns higher weights to frames that are closer in time to the central frame and lower weights to frames that are farther away. By taking the weighted average of these frames, we obtain a single, representative apex frame. Fig. 1(a) represents an example of the generation of apex frame. This procedure is designed to extract the most significant details from the video sequence, compressing them into a single frame for subsequent analysis. Motivated by this, we generate unlabeled data by considering different temporal lengths of the video as shown in Fig. 1(b) . To combine this complementary information, we present a semi-supervised learning to use both labeled and unlabeled data to improve the performance of face presentation attack detection. In summary, our key contributions can be summarized as follow:

\begin{enumerate}
 \item This study presents the first attempt to employ the apex frame for the summarization of an entire video into a singular image, with a specific focus on enhancing facial anti-spoofing task.
\item  In contrast to frame-based approaches (i.e., optical flow or global motion), our approach is computationally faster since it does not involve estimating pixel-level motion vectors between frames. This computational efficiency makes our method more practical and scalable for real-world applications.
\item Through semi-supervised learning, we evaluated the effectiveness of our approach on four different datasets. The results demonstrate promising generalization ability, achieving new state-of-the-art performance on three datasets.
\end{enumerate}
The paper is structured as follows: Section 2 explains the proposed methodology, Section 3 discusses experimental settings and comparison results, and Section 4 presents conclusions.

\section{Methodology}
\label{sec:majhead}
The proposed approach comprises three key steps: (1) creating an apex frame, (2) generating unlabeled data, and (3) implementing semi-supervised learning. Fig. 2 provides an overall visualization of the proposed method, illustrating how the labeled and unlabeled data is interconnected through semi-supervised learning to address the challenge of face presentation attack detection.

\subsection{Apex frame creation}
The selection of important frames for face anti-spoofing is a relatively recent focus in biometric authentication research. This is primarily due to the high degree of similarity in spatial and motion information found in many neighboring frames \cite{muhammad2023deep}. These frames, whether derived from genuine or attack videos, often provide less significant structural and textural information to the overall analysis. Taking inspiration from Gaussian distribution theory, we employ a Gaussian distribution function to generate an apex frame by calculating Gaussian weights, with a focus on the central frame of the video. The standard deviation determines the spread or width of the Gaussian distribution, affecting the weights assigned to frames in the apex frame generation process. We implement our approach by employing the following mathematical equations.

Suppose $I_i$ is the $i$-th frame in the video and $I_c$ is the central frame. The Gaussian weighting for frame $I_i$ relative to the central frame $I_c$ can be calculated as follows:
\begin{equation} \label{eq1}
W_i = \exp\left(-\frac{(i - c)^2}{2\sigma^2}\right),
\end{equation}
where $W_i$ is the weight assigned to frame $I_i$, $i$ is the index of frame $I_i$ and  $c$ is the index of the central frame $I_c$. $\sigma$ is the standard deviation parameter that controls the spread of the Gaussian distribution. 

After calculating the weights for all frames in the video, we normalize them as follow:
\begin{equation} \label{eq2}
\text{Normalized Weight}_i = \frac{W_i}{\sum_{i=1}^{N} W_i},
\end{equation}
where $N$ is the total number of frames in the video. Finally, to compute the apex frame $F$ , we use the following equation:
\begin{equation} \label{eq3}
F = \sum_{i=1}^{N} \left(\text{Normalized Weight}_i \cdot I_i\right).
\end{equation}

This single weighted apex frame gives a representation of the video that focuses more on frames that are located closer to the central frame $I_c$, while frames farther away have lower influence due to the Gaussian weighting.  Thus, generating apex frames reduces facial data redundancy and provides the most informative and representative of the overall content. 

\begin{figure}[]
  \centering
  \centerline{\includegraphics[width=8.2cm]{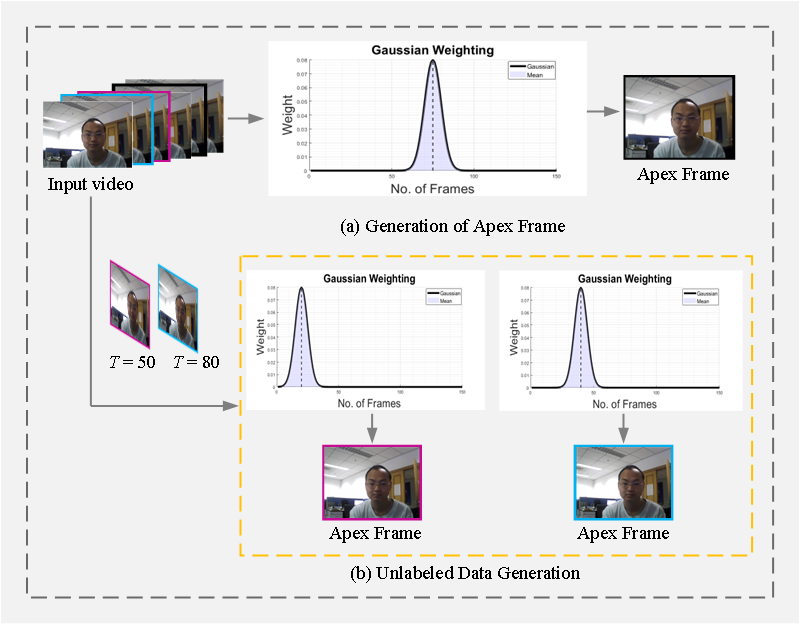}}
 \caption{The flowchart of the proposed Gaussian weighted apex frame method.
(a) The Gaussian weighted curve indicates how much influence or weight is assigned to frames at different temporal distances from the central frame. (b)  Unlabeled data is generated by utilizing various temporal lengths of the video centered around the apex (central) frame of the segment.}
\end{figure}

\begin{figure*}[htb]
  \centering
  \centerline{\includegraphics[width=11.9cm]{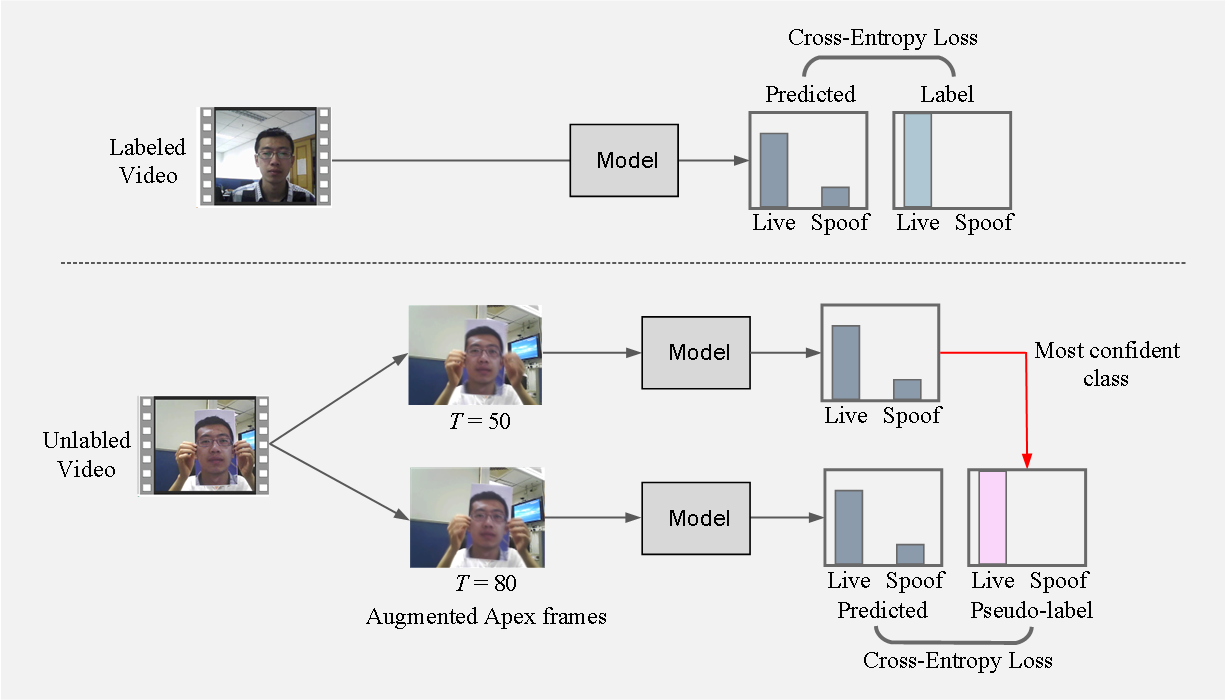}}
 \caption{A schematic diagram of our proposed semi-supervised learning.}
\end{figure*}

\subsection{Unlabeled data generation}
The primary concept behind our unlabeled data generation involves splitting video clips into different temporal lengths denoted as $T$, and then using this pseudo data to learn the  spatiotemporal variations and context in an unsupervised manner. In particular, the model does not rely on explicit labels or annotations. Instead, it discover patterns and structure in the data without human-provided guidance. To achieve this, we divide the video into segments. Each segment, denoted as $S_j$, contains a subset of $N=50$ frames from the video. For each segment $S_j$, we calculate Gaussian weights for the frames within that segment relative to the central frame $I_{c_j}$ of that segment. The Gaussian weighting for frame $I_i^j$ within segment $S_j$ is given by:

\begin{equation} \label{eq3}
W_i^j = \exp\left(-\frac{(i - c_j)^2}{2\sigma^2}\right),
\end{equation}
where $W_i^j$ is the weight assigned to frame $I_i^j$ in segment $S_j$, $i$ is the index of frame $I_i^j$ within segment $S_j$, $c_j$ is the index of the central frame $I_{c_j}$ in segment $S_j$ and  $\sigma$ is the standard deviation parameter that controls the spread of the Gaussian distribution. 

After calculating the Gaussian weights for frames within each segment, we create a weighted frame $F_j$ that represents a single apex frame for segment $S_j$ by taking the weighted sum of frames:

\begin{equation} \label{eq3}
F_j = \sum_{i=1}^{N_j} \left(\frac{W_i^j}{\sum_{k=1}^{N_j} W_k^j} \cdot I_i^j\right),
\end{equation}
where $N_j$ is the total number of frames in segment $S_j$. $W_i^j$ is the Gaussian weight for frame $I_i^j$ in segment $S_j$. $I_i^j$ is the $i$-th frame in segment $S_j$. We iterate through this process, considering each successive video segment to obtain distinct apex frames for various sections of the video.

\subsection{Semi-Supervised learning}
Our approach leverages the benefits of having access to labeled data obtained from Eq. 3 while also utilizing the large pool of unlabeled data extracted from Eq. 5 to improve model performance. In particular, we generate two pseudo classes of Gaussian-guided outputs with distinct temporal lengths (T=50 and T=80). These pseudo categories are used to initialize a semi-supervised learning process. By doing so, the model undergoes typical supervised training with labeled images, utilizing a cross-entropy loss function. Subsequently, the same model is employed to generate predictions for a batch of unlabeled images, selecting the class with the highest confidence as the pseudo-label. Then, a cross-entropy loss is computed by comparing the model's predictions with the pseudo-labels assigned to the unlabeled images. The total loss for training is a combination of the supervised loss and the unsupervised loss, and utilized as:
\begin{equation} \label{eq3}
\text{L} = L_{\textit{labeled}} + \lambda \cdot L_{\textit{unlabeled}},
\end{equation}
where $\lambda$ is a hyperparameter. In the proposed semi-supervised learning approach, we employ a 2D convolutional neural network \cite{huang2017densely} while also leveraging the capabilities of long short-term memory (LSTM) network for temporal modeling \cite{hochreiter1997long}.

\begin{table*}[t]
\centering
\caption{Performance evaluation using MSU-MFSD (M), Idiap (I), CASIA (C) and OULU-NPU (O) databases. Comparison results are obtained directly from the cited papers.} \label{tab:cap2}
\vspace{7pt} 
\begin{tabular}{l|r|l|r | l| r | l|r| l}
\hline
 & \multicolumn{2}{c|}{O\&C\&I to M} & \multicolumn{2}{c|}{O\&M\&I to C}  & \multicolumn{2}{c|}{O\&C\&M to I} & \multicolumn{2}{c}{I\&C\&M to O} \\ \hline
  Method     & HTER   & AUC   & HTER   & AUC  & HTER    & AUC    & HTER  & AUC   \\ \hline 
MADDG  \cite{shao2019multi} & 17.69    & 88.06   &  24.50 &  84.51 & 22.19 &   84.99 &  27.89 &   80.02 \\
DAFL  \cite{saha2020domain} & 14.58    & 92.58   &  17.41 &  90.12 & 15.13 &  95.76 &  14.72 &  93.08 \\ 
DR-MD  \cite{wang2020cross} & 17.02   & 90.10 &  19.68 &  87.43 & 20.87 & 86.72 &  25.02 &  81.47 \\
GW + RNN  \cite{muhammad2023face} & \underline{4.12}    &  \underline {99.93}  &  \underline{7.04} & \textbf {99.87}  & 13.48 & 97.42  &  41.33 &  88.48  \\
RFMetaFAS  \cite{shao2020regularized} & 13.89    &  93.98  &  20.27 &  88.16 & 17.30 & 90.48 &  16.45 &  91.16 \\
FAS-DR-BC(MT)  \cite{qin2021meta} & 11.67    & 93.09   &  18.44 &  89.67 & 11.93 & 94.95 &  16.23 &  91.18 \\
HFN + MP  \cite{cai2022learning} & 5.24 & 97.28 &  9.11  &   96.09 &   15.35 &   90.67  & \underline{12.40} &  94.26 \\ 
Cross-ADD \cite{huang2022generalized}  & 11.64   & 95.27 &   17.51  & 89.98  &  15.08 &  91.92  &  14.27  &    93.04  \\ 
FG +HV  \cite{liu2022feature} & 9.17   & 96.92  &  12.47 &  93.47 & 16.29  &  90.11 &  13.58 &  93.55 \\
ADL  \cite{liu2022adversarial} &   5.00   &   97.58    & 10.00 &  96.85 & 12.07 &   94.68  & 13.45  &  94.43 \\ 
Regression Network  \cite{kwak2023liveness} &   5.41    &   98.85   & 10.05 &  94.27 & 8.62 &   \underline{97.60} &  \textbf {11.42} &  \underline{95.52} \\ 
CRFAS \cite{zheng2023learning} &   7.14   &   97.44  & 9.88 &  96.56 & \underline{8.57} &   96.07 & 16.38 &  90.87  \\ 
\hline 
Supervised learning  &   10.41   &   96.21    & 26.11 &  82.28 & 23.12 &   90.38  & 45.99  &  81.12 \\ 
Semi-supervised learning &   7.08   &   98.27    & 15.74 &  90.97 & 21.62 &   92.65  & 38.97  &  79.64 \\ 
           
Semi-supervised + LSTM &  \textbf {3.10} & \textbf {99.95}  & \textbf {4.12} &  \underline{98.49}  &   \textbf {7.37} & \textbf {99.18}  &  28.92 & \textbf {96.44} \\ 
\hline\end{tabular}
\end{table*} 

\begin{figure*}
   \centering
       \begin{subfigure}[b]{0.24\textwidth}
        \centering
          \includegraphics[width=\textwidth]{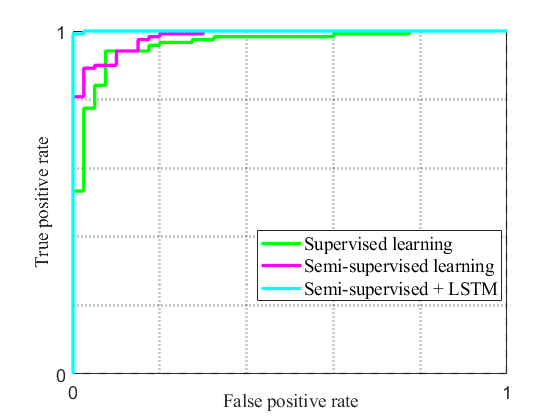}
         \caption{}
     \end{subfigure}
     \hfill
     \begin{subfigure}[b]{0.24\textwidth}
        \centering
          \includegraphics[width=\textwidth]{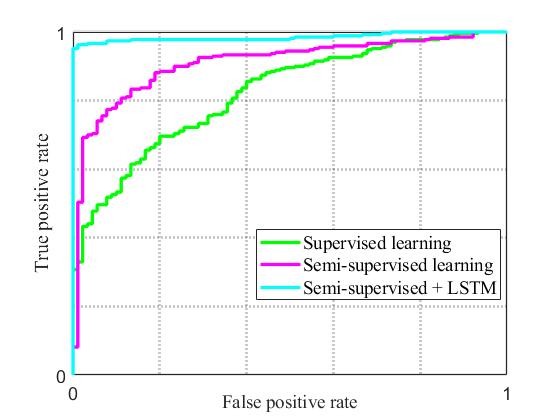}
         \caption{}
     \end{subfigure}
     \hfill
     \begin{subfigure}[b]{0.24\textwidth}
        \centering
         \includegraphics[width=\textwidth]{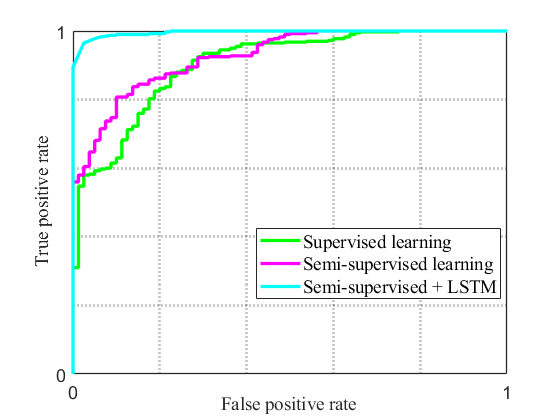}
         \caption{}
     \end{subfigure}
      \hfill
     \begin{subfigure}[b]{0.24\textwidth}
        \centering
          \includegraphics[width=\textwidth]{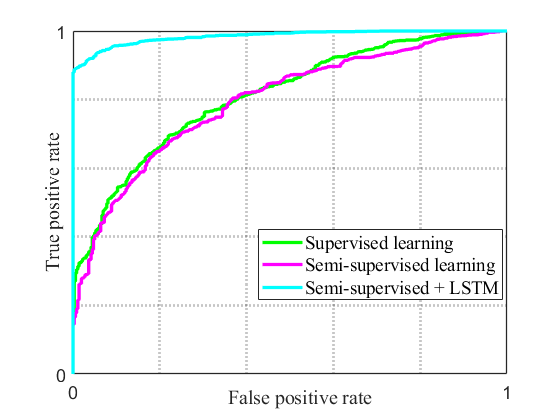}
         \caption{}
     \end{subfigure}
          \hfill
        \caption{The Receiver Operating Characteristics (ROC) curves. (a) O\&C\&I to M, (b) O\&M\&I to C, (c) O\&C\&M to I, and (d) I\&C\&M to O illustrates ROC curves for four datasets.}
\end{figure*}

\section{Experiments}
\label{sec:experiments}
To evaluate the effectiveness of our proposed method, our study employs several databases: Idiap Replay-attack \cite{chingovska2012effectiveness} (denoted as I), MSU Mobile Face Spoofing \cite{wen2015face} (denoted as M), OULU-NPU \cite{boulkenafet2017oulu} (denoted as O), and CASIA Face Anti-Spoofing \cite{zhang2012face} (referred to as C). We report the Half Total Error Rate (HTER), taking into consideration both false acceptances and false rejections. We also define the Area Under the Curve (AUC) as an inclusive metric that provides a singular numerical value summarizing the overall performance of the model.

\subsection{Implementation details}
The Densely Connected Convolutional Networks (DenseNet-201) \cite{huang2017densely} model is fine-tuned on the labeled data by resizing all images to $224 \times 224$. Traditional data augmentation transformations such as rotation, x-translation, and y-translation were performed to improve the robustness of the model. During the fine-tuning process, we employed specific parameters, including the Adam optimizer, a validation frequency of every $30$ steps, and a learning rate set at $0.0001$ with an early stopping function for model development. The $\lambda$ cofficient is set to $1.5$. To generate the apex frame and produce unlabeled data, we consistently use standard deviation $\sigma=5$. We set a confidence threshold $0.9$ for the semi-supervised learning. This expanded training dataset includes unlabeled data with reliably predicted labels.

For training the LSTM, the deep features , comprising $1920$ dimensions, were extracted directly from the last pooling layer of the model after fine-tuned the model in semi-supervised way. The training process of LSTM involves the Adam optimizer, a hidden layer dimension of $100$, a validation frequency of $30$, and a learning rate of $0.0001$ with an early stopping function. We use the same parameters for training across all datasets to ensure the repeatability of our results.

\subsection{Comparison against state-of-the-art methods}
To assess the efficacy of our proposed approach, we conducted a comparison with various state-of-the-art deep learning models as shown in Table 1. In particular, we employed a leave-one-out (LOO) strategy, in which three datasets were randomly chosen for training, and one dataset was held out for testing in each experiment. Among recently reported domain generalization methods \cite{kwak2023liveness, zheng2023learning, cai2022learning, yu2020fas, saha2020domain, shao2019multi, muhammad2022adaptive, cai2022learning, liu2022adversarial}, we noted that the proposed semi-supervised learning provides state-of-the-art performance on three testing sets (i.e., MSU, CASIA, and Replay-attack). It is worth noting that the equal error rate is calculated using source testing sets, and subsequently, a threshold is applied to determine the Half Total Error Rate (HTER) on an unseen target testing set. We hypothesis that Gaussian guided apex frame can enhance the model's generalization by focusing on the most informative moments of a video. When the model learns from unlabeled Apex frames, it becomes better at recognizing essential patterns and content in videos, making it more versatile in handling unseen video data. 

Additionally, as demonstrated in Table 1, it is evident that our proposed method consistently achieves AUC scores exceeding  $90\%$ across all the datasets. For further analysis, the ROC curves are visualized in Fig. 3  by plotting the true positive rate (TPR) on the y-axis and the false positive rate (FPR) on the x-axis. One can observe at the ROC curves that they are positioned in the top-left corner. This positioning signifies higher true positive rates and lower false positive rates, which in turn demonstrate the model's improved ability to discriminate effectively in comparison to the model trained without unlabeled data.

\section{Conclusions}
 In this paper, we highlight the significance of apex frame in the face anti-spoofing task. Leveraging Gaussian weighting for apex frame selection allows the model to efficiently concentrate on the most relevant areas. Our study reveals that apex frames with varying temporal lengths complement semi-supervised learning, achieving state-of-the-art performance across three previously unseen face anti-spoofing datasets. Furthermore, training the model on a more compact and informative apex frames enhances its generalization to unseen data. However, the paper acknowledges a potential limitation: the model may not capture all fine details, especially in longer videos. Therefore, future research should explore new approaches and enhancements to address this limitation and advance video summarization methods for longer videos.




\bibliographystyle{IEEEbib}
\bibliography{strings,refs}

\begin{thebibliography}{10}

\bibitem{muhammad2023self}
Usman Muhammad and Mourad Oussalah,
\newblock ``Self-supervised face presentation attack detection with dynamic grayscale snippets,''
\newblock in {\em 2023 IEEE 17th International Conference on Automatic Face and Gesture Recognition (FG)}. IEEE, 2023, pp. 1--6.

\bibitem{muhammad2023deep}
Usman Muhammad, Md~Ziaul Hoque, Mourad Oussalah, and Jorma Laaksonen,
\newblock ``Deep ensemble learning with frame skipping for face anti-spoofing,''
\newblock {\em arXiv preprint arXiv:2307.02858}, 2023.

\bibitem{shao2020regularized}
Rui Shao, Xiangyuan Lan, and Pong~C Yuen,
\newblock ``Regularized fine-grained meta face anti-spoofing.,''
\newblock in {\em AAAI}, 2020, pp. 11974--11981.

\bibitem{muhammad2023domain}
Usman Muhammad, Djamila~Romaissa Beddiar, and Mourad Oussalah,
\newblock ``Domain generalization via ensemble stacking for face presentation attack detection,''
\newblock {\em arXiv preprint arXiv:2301.02145}, 2023.

\bibitem{muhammad2023saliency}
Usman Muhammad, Mourad Oussalah, Md~Ziaul Hoque, and Jorma Laaksonen,
\newblock ``Saliency-based video summarization for face anti-spoofing,''
\newblock {\em arXiv preprint arXiv:2308.12364}, 2023.

\bibitem{wang2020cross}
Guoqing Wang, Hu~Han, Shiguang Shan, and Xilin Chen,
\newblock ``Cross-domain face presentation attack detection via multi-domain disentangled representation learning,''
\newblock in {\em CVPR}, 2020, pp. 6678--6687.

\bibitem{muhammad2019faceb}
Usman Muhammad, Tuomas Holmberg, Wheidima Carneiro~de Melo, and Abdenour Hadid,
\newblock ``Face anti-spoofing via sample learning based recurrent neural network (rnn),''
\newblock in {\em The British Machine Vision Conference 2019 (BMVC) 9th-12th September 2019, Cardiff UK}. British Machine Vision Association Press, 2019.

\bibitem{boulkenafet2016face}
Zinelabidine Boulkenafet, Jukka Komulainen, and Abdenour Hadid,
\newblock ``Face spoofing detection using colour texture analysis,''
\newblock {\em TIFS}, vol. 11, no. 8, pp. 1818--1830, 2016.

\bibitem{muhammad2019face}
Usman Muhammad and Abdenour Hadid,
\newblock ``Face anti-spoofing using hybrid residual learning framework,''
\newblock in {\em 2019 International Conference on Biometrics (ICB)}. IEEE, 2019, pp. 1--7.

\bibitem{wen2015face}
Di~Wen, Hu~Han, and Anil~K Jain,
\newblock ``Face spoof detection with image distortion analysis,''
\newblock {\em TIFS}, vol. 10, no. 4, pp. 746--761, 2015.

\bibitem{siddiqui2016face}
Talha~Ahmad Siddiqui, Samarth Bharadwaj, Tejas~I Dhamecha, Akshay Agarwal, Mayank Vatsa, Richa Singh, and Nalini Ratha,
\newblock ``Face anti-spoofing with multifeature videolet aggregation,''
\newblock in {\em 2016 23rd International Conference on Pattern Recognition (ICPR)}. IEEE, 2016, pp. 1035--1040.

\bibitem{benlamoudi2017face}
Azeddine Benlamoudi, Kamal~Eddine Aiadi, Abdelkrim Ouafi, Djamel Samai, and Mourad Oussalah,
\newblock ``Face antispoofing based on frame difference and multilevel representation,''
\newblock {\em Journal of Electronic Imaging}, vol. 26, no. 4, pp. 043007--043007, 2017.

\bibitem{muhammad2022self}
Usman Muhammad, Zitong Yu, and Jukka Komulainen,
\newblock ``Self-supervised 2d face presentation attack detection via temporal sequence sampling,''
\newblock {\em Pattern Recognition Letters}, vol. 156, pp. 15--22, 2022.

\bibitem{muhammad2023face}
Usman Muhammad and Mourad Oussalah,
\newblock ``Face anti-spoofing from the perspective of data sampling,''
\newblock {\em Electronics Letters}, vol. 59, no. 1, pp. e12692, 2023.

\bibitem{li2020joint}
Yante Li, Xiaohua Huang, and Guoying Zhao,
\newblock ``Joint local and global information learning with single apex frame detection for micro-expression recognition,''
\newblock {\em IEEE Transactions on Image Processing}, vol. 30, pp. 249--263, 2020.

\bibitem{huang2017densely}
Gao Huang, Zhuang Liu, Laurens Van Der~Maaten, and Kilian~Q Weinberger,
\newblock ``Densely connected convolutional networks,''
\newblock in {\em Proceedings of the IEEE conference on computer vision and pattern recognition}, 2017, pp. 4700--4708.

\bibitem{hochreiter1997long}
Sepp Hochreiter and J{\"u}rgen Schmidhuber,
\newblock ``Long short-term memory,''
\newblock {\em Neural computation}, vol. 9, no. 8, pp. 1735--1780, 1997.

\bibitem{shao2019multi}
Rui Shao, Xiangyuan Lan, Jiawei Li, and Pong~C Yuen,
\newblock ``Multi-adversarial discriminative deep domain generalization for face presentation attack detection,''
\newblock in {\em CVPR}, 2019, pp. 10023--10031.

\bibitem{saha2020domain}
Suman Saha, Wenhao Xu, Menelaos Kanakis, Stamatios Georgoulis, Yuhua Chen, Danda~Pani Paudel, and Luc Van~Gool,
\newblock ``Domain agnostic feature learning for image and video based face anti-spoofing,''
\newblock in {\em Proceedings of the IEEE Conference on Computer Vision and Pattern Recognition Workshops}, 2020, pp. 802--803.

\bibitem{qin2021meta}
Yunxiao Qin, Zitong Yu, Longbin Yan, Zezheng Wang, Chenxu Zhao, and Zhen Lei,
\newblock ``Meta-teacher for face anti-spoofing,''
\newblock {\em IEEE transactions on pattern analysis and machine intelligence}, vol. 44, no. 10, pp. 6311--6326, 2021.

\bibitem{cai2022learning}
Rizhao Cai, Zhi Li, Renjie Wan, Haoliang Li, Yongjian Hu, and Alex~C Kot,
\newblock ``Learning meta pattern for face anti-spoofing,''
\newblock {\em IEEE Transactions on Information Forensics and Security}, vol. 17, pp. 1201--1213, 2022.

\bibitem{huang2022generalized}
Hanye Huang, Youjun Xiang, Guodong Yang, Lingling Lv, Xianfeng Li, Zichun Weng, and Yuli Fu,
\newblock ``Generalized face anti-spoofing via cross-adversarial disentanglement with mixing augmentation,''
\newblock in {\em ICASSP 2022-2022 IEEE International Conference on Acoustics, Speech and Signal Processing (ICASSP)}. IEEE, 2022, pp. 2939--2943.

\bibitem{liu2022feature}
Shice Liu, Shitao Lu, Hongyi Xu, Jing Yang, Shouhong Ding, and Lizhuang Ma,
\newblock ``Feature generation and hypothesis verification for reliable face anti-spoofing,''
\newblock in {\em Proceedings of the AAAI Conference on Artificial Intelligence}, 2022, pp. 1782--1791.

\bibitem{liu2022adversarial}
Mingxin Liu, Jiong Mu, Zitong Yu, Kun Ruan, Baiyi Shu, and Jie Yang,
\newblock ``Adversarial learning and decomposition-based domain generalization for face anti-spoofing,''
\newblock {\em Pattern Recognition Letters}, vol. 155, pp. 171--177, 2022.

\bibitem{kwak2023liveness}
Youngjun Kwak, Minyoung Jung, Hunjae Yoo, JinHo Shin, and Changick Kim,
\newblock ``Liveness score-based regression neural networks for face anti-spoofing,''
\newblock in {\em ICASSP 2023-2023 IEEE International Conference on Acoustics, Speech and Signal Processing (ICASSP)}. IEEE, 2023, pp. 1--5.

\bibitem{zheng2023learning}
Guanghao Zheng, Yuchen Liu, Wenrui Dai, Chenglin Li, Junni Zou, and Hongkai Xiong,
\newblock ``Learning causal representations for generalizable face anti spoofing,''
\newblock in {\em ICASSP 2023-2023 IEEE International Conference on Acoustics, Speech and Signal Processing (ICASSP)}. IEEE, 2023, pp. 1--5.

\bibitem{chingovska2012effectiveness}
Ivana Chingovska, Andr{\'e} Anjos, and S{\'e}bastien Marcel,
\newblock ``On the effectiveness of local binary patterns in face anti-spoofing,''
\newblock in {\em Proceedings of the international conference of biometrics special interest group}. IEEE, 2012, pp. 1--7.

\bibitem{boulkenafet2017oulu}
Zinelabinde Boulkenafet, Jukka Komulainen, Lei Li, Xiaoyi Feng, and Abdenour Hadid,
\newblock ``Oulu-npu: A mobile face presentation attack database with real-world variations,''
\newblock in {\em FG}. IEEE, 2017, pp. 612--618.

\bibitem{zhang2012face}
Zhiwei Zhang, Junjie Yan, Sifei Liu, Zhen Lei, Dong Yi, and Stan~Z Li,
\newblock ``A face antispoofing database with diverse attacks,''
\newblock in {\em 2012 5th IAPR international conference on Biometrics (ICB)}. IEEE, 2012, pp. 26--31.

\bibitem{yu2020fas}
Zitong Yu, Jun Wan, Yunxiao Qin, Xiaobai Li, Stan~Z Li, and Guoying Zhao,
\newblock ``Nas-fas: Static-dynamic central difference network search for face anti-spoofing,''
\newblock {\em IEEE transactions on pattern analysis and machine intelligence}, vol. 43, no. 9, pp. 3005--3023, 2020.

\bibitem{muhammad2022adaptive}
Usman Muhammad, Jiehua Zhang, Li~Liu, and Mourad Oussalah,
\newblock ``An adaptive spatio-temporal global sampling for presentation attack detection,''
\newblock {\em IEEE Transactions on Circuits and Systems II: Express Briefs}, 2022.

\end{thebibliography}

\end{document}